\documentclass[runningheads]{llncs}
\usepackage{graphicx}
\usepackage{subfigure}
\usepackage{amsmath}
\usepackage{algorithmic}
\usepackage{algorithm}

\usepackage{xcolor}

\definecolor{Yellow}{rgb}{1,0.9,0.7}
\definecolor{Pink}{rgb}{1,0.85,0.85}
\definecolor{AntiqueWhite}{rgb}{0.9,0.9,0.9}

\newcommand{\NOTE}[1]%
{
\noindent
\fboxsep=2mm\fcolorbox{black}{AntiqueWhite}{\parbox{0.95\columnwidth}
{\textbf{NOTE: } #1}
}
}

\begin{document}
\title{Asynchronous Corner Tracking Algorithm based on Lifetime of Events for DAVIS Cameras\thanks{Supported by the Academy of Finland under the project (314048)}}
\titlerunning{Asynchronous Corner Tracking Algorithm}
%
\author{Sherif A.S. Mohamed\inst{1}\and
Jawad N. Yasin\inst{1}\and Mohammad-Hashem~Haghbayan\inst{1}\and{Antonio Miele}\inst{3} \and Jukka Heikkonen\inst{1}\and Hannu Tenhunen\inst{2}\and Juha Plosila\inst{1}}
\authorrunning{Sherif A.S. Mohamed et al.}
%
\institute{University of Turku, 20500 Turku, Finland \and
KTH Royal Institute of Technology, 11428 Stockholm, Sweden \and Politecnico di Milano, 20133 Milano, Italy}
\maketitle              
\begin{abstract}
Event cameras, i.e., the Dynamic and Active-pixel Vision Sensor (DAVIS) ones, capture the intensity changes in the scene and generates a stream of events in an asynchronous fashion. The output rate of such cameras can reach up to 10 million events per second in high dynamic environments. DAVIS cameras use novel vision sensors that mimic human eyes. Their attractive attributes, such as high output rate, High Dynamic Range (HDR), and high pixel bandwidth, make them an ideal solution for applications that require high-frequency tracking. Moreover, applications that operate in challenging lighting scenarios can exploit from the high HDR of event cameras, i.e., 140 dB compared to 60 dB of traditional cameras. In this paper, a novel asynchronous corner tracking method is proposed that uses both events and intensity images captured by a DAVIS camera. The Harris algorithm is used to extract features, i.e., frame-corners from keyframes, i.e., intensity images. Afterward, a matching algorithm is used to extract event-corners from the stream of events. Events are solely used to perform asynchronous tracking until the next keyframe is captured. Neighboring events, within a window size of 5x5 pixels around the event-corner, are used to calculate the velocity and direction of extracted event-corners by fitting the 2D planar using a randomized Hough transform algorithm. Experimental evaluation showed that our approach is able to update the location of the extracted corners up to 100 times during the blind time of traditional cameras, i.e., between two consecutive intensity images.

\end{abstract}
\section{Introduction}
Simultaneous Localization And Mapping (SLAM) approaches use onboard sensors, such as Lidar, camera, and radar to observe the environment and estimate the position and the orientation of the robot \cite{sherif} \cite{jd1} \cite{jd2}. For example, visual SLAM methods use single or multiple cameras to obtain the pose by extracting and tracking a sufficient number of features from consecutive images. Over the past years, there are many methods have been presented to detect frame-corners (e.g., \textit{Harris} \cite{harris}) and edges (e.g., \textit{Canny}~\cite{canny}).

Most of SLAM approaches in the literature obtain the pose of robots by progressing a sequence of intensity images captured by CMOS sensors, i.e., frame-based cameras. However, such cameras mostly suffer from some major hardware limitations. They generate grayscale or RGB images at a fixed rate, typically 60Hz. Moreover, they produce unnecessary information when the camera is not moving and the scene is static, which increases the computational cost dramatically. On the other hand, when the camera and objects in the scene are highly dynamic they are affected by the motion blur phenomena, i.e., they generate insufficient information. In addition, a huge amount of information might be missed during the blind time, i.e., the time between two successive images, which might result in inaccurate pose estimates. In conclusion, the limitations of conventional cameras would degrade the tracking performance, resulting in inaccurate localization. 

\begin{figure}[t]
\centering
\subfigure[The camera is moving rapidly]{\includegraphics[width = 2.1in]{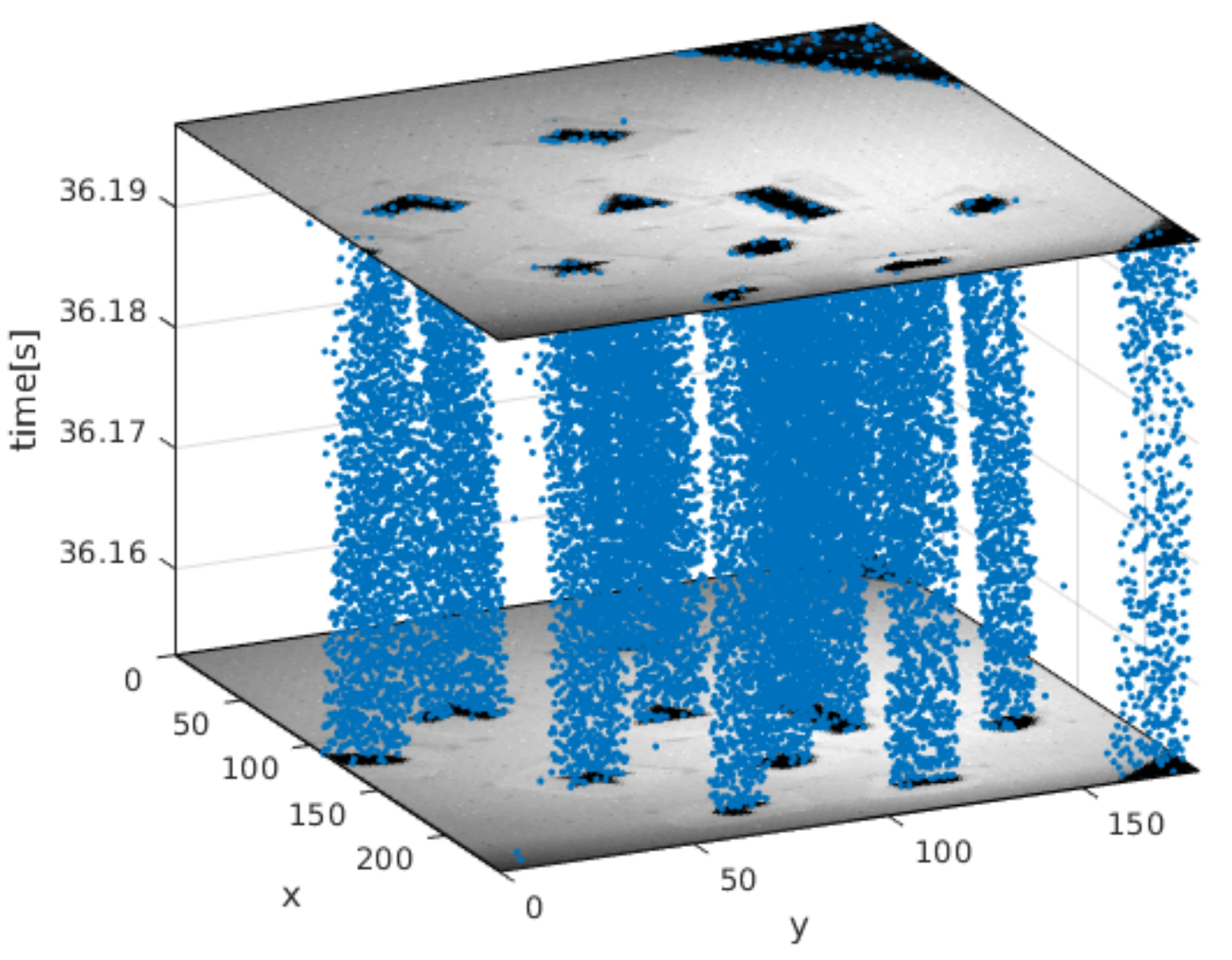}\label{fig:event1}} \hspace{2pt}
\subfigure[The camera and scene are static]{\includegraphics[width = 2.1in]{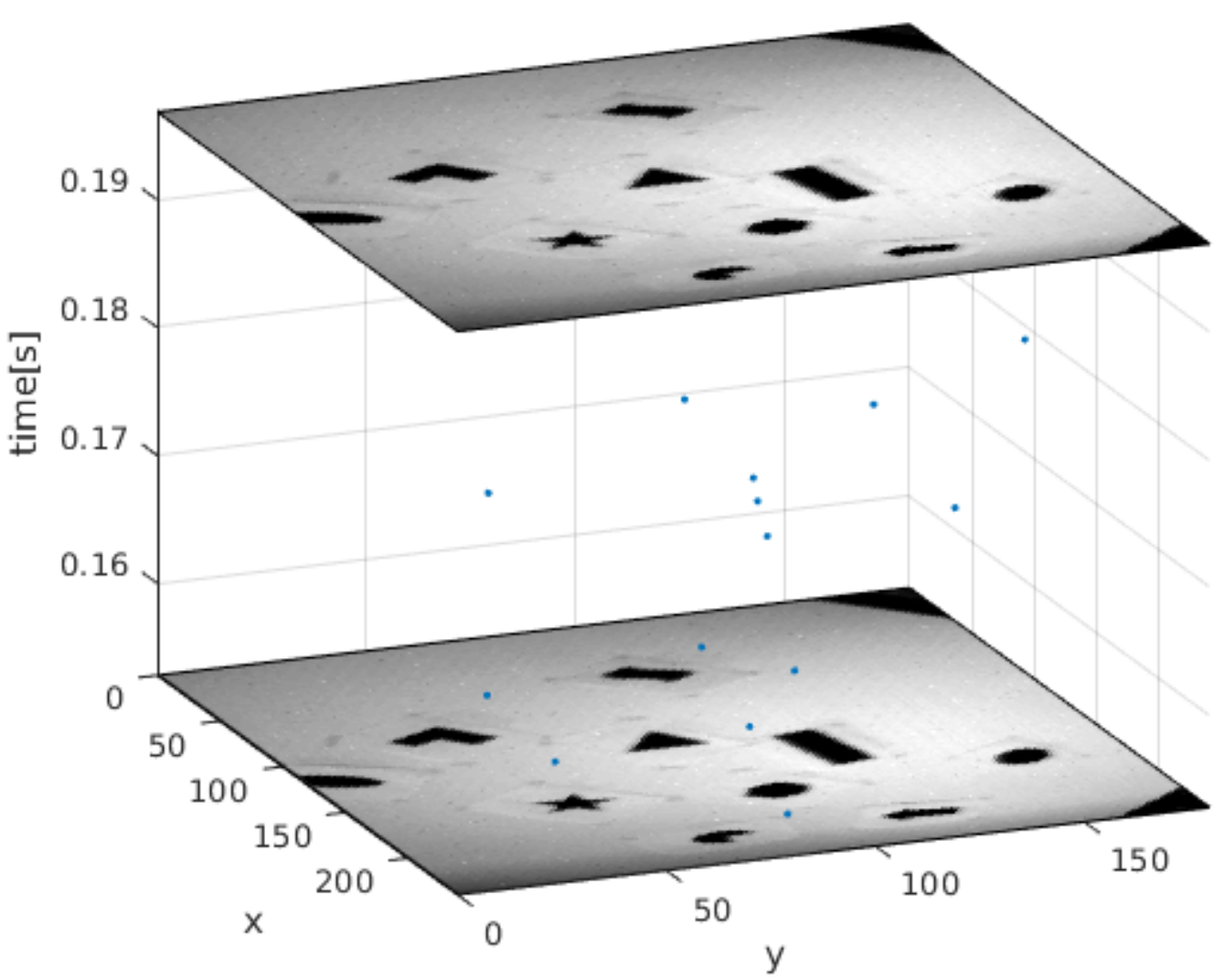}\label{fig:event2}}

\caption{Events (blue) triggered between two consecutive frame-images in two different scenarios.}
\label{fig:event}
\end{figure}

Newly emerged event cameras, such as DVS \cite{dvs} and ATIS \cite{atis}, overcome these problems through transmitting a stream of events by capturing the intensity changes instead of the ``absolute'' intensity. Events are triggered when the intensity of any pixel changes by a certain threshold. Thus, the number of triggered events depends on the texture of the environment and the motion of the camera and objects in the scene. rightness increases ``1'' or decreases ``0''.  Figure \ref{fig:event} illustrates the output of the event camera in two different.

When the camera and the scene are static or the camera is moving slowly in a texture-less scene, event cameras only generate a few numbers of events and therefore only a small amount of resources would be enough to process those events (see Figure \ref{fig:event2}). Contrarily, when the scene is highly dynamic or the camera is moving rapidly, a large number of events (in order of millions) will be generated (see Figure \ref{fig:event1}). These powerful attributes make event cameras a great solution for applications that operates in high dynamic scenes and need to get their pose updated frequently, i.e., drones. 

Given to the discussed necessity of robust and high-frequency corner detection algorithms, in this paper we present a novel algorithm that exploits the attractive attributes of event cameras and operates in an asynchronous fashion. Our algorithm uses both events and grayscale images captured by the DAVIS \cite{davis} cameras to extract and track corners. The proposed algorithm is composed of three main phases: observation, detection, and asynchronous tracking. A sufficient number of salient frame-corners are detected using Harris \cite{harris} from grayscale images. Afterward, a matching filter is applied to obtain event-corners by selecting the first event that occurs on the same pixel location of extracted frame-corners. A window size of 5x5 pixels around each event-corner is used to compute the lifetime of each event-corner. We use a randomized Hough transform to fit a local plane to the points in the 5x5 pixels matrix~\cite{RHT} and compute the lifetime and the direction of each event-corner. Lifetime indicates the time a event-corner will take to move from its current pixel to one of the eight neighboring pixels. Our algorithm is able to update the extracted corners up to 2.4$\times 10^3$ times per second compared to 24 using only images. 

The rest of the paper is organized as follows.
Section 2 reviews related works In Section 3, the system architecture of the proposed work is presented. An experimental evaluation of the proposed approach is discussed in Section 4. Last section draws conclusion and presents future work.

\section{Related Work}

To unlock  events' camera potential, novel methods are required to process the output, i.e., a stream of events. Methods in the literature can be categorized as indirect and direct. 

Indirect methods pre-process the captured events to generate artificially synthesized frames, i.e., event-frames and then apply one of the state-of-the-art frame-based methods to extract and track frame-corners. In \cite{deltaT}, the authors generate event-frames by accumulating events during a fixed time interval. In \cite{evo}, the authors proposed another approach to generate event-frames by accumulating a fixed number of events, (e.g., 2000 events) to form a frame. In \cite{sherif1}, the authors presented a dynamic slicing method to generate event-frames based on the velocity of the event camera and the number of edges in the environment, i.e., \textit{entropy}. The main disadvantage of using such techniques is that they omit the innate asynchronous nature of the event cameras. 

Direct approaches, on the other hand, process the asynchronous stream of events directly. In \cite{eharris}, Vasco et al. proposed an algorithm to detect asynchronous corners using an adaptation of the original Harris score \cite{harris} on a spatio-temporal window, i.e., surface Active Events (SAE). The algorithm demands a lot of computational resources to compute the gradients of each incoming event. In \cite{eFAST}, the authors present an algorithm inspired by FAST \cite{fast} to detect event-corners, they call it eFAST. Corners are extracted on SAE by comparing the timestamp of the latest events, i.e., pixels on two circles. The radius of the inner and the outer circle is 3 and 4 pixels respectively. In \cite{arc}, the authors used a similar technique (Arc) with an event filter to detect corner corners 4.5x faster than the eFAST. There are two main disadvantages of using such techniques mainly related to computational cost and accuracy. Since event cameras can generate millions of events per second and processing each incoming event to extract corners is computationally expensive. Thus, they are not feasible to run in real-time performance and especially on resource contained systems. On the other hand, some approaches rely on simple operations to perform real-time corner extraction. However, these methods produce inaccurate corners $(50 \sim 60 \%)$ compared to frame-based methods, such as Harris.

In conclusion, indirect methods omit one of the main characteristics of event-cameras, i.e., asynchronous. Direct methods, on the other hand, mistakenly detect more false event-corners and subsequently require a lot of computational resources to process all the incoming events. 

\section{Proposed approach}

\begin{figure}[t]
    \centering
    \includegraphics[width=1\textwidth]{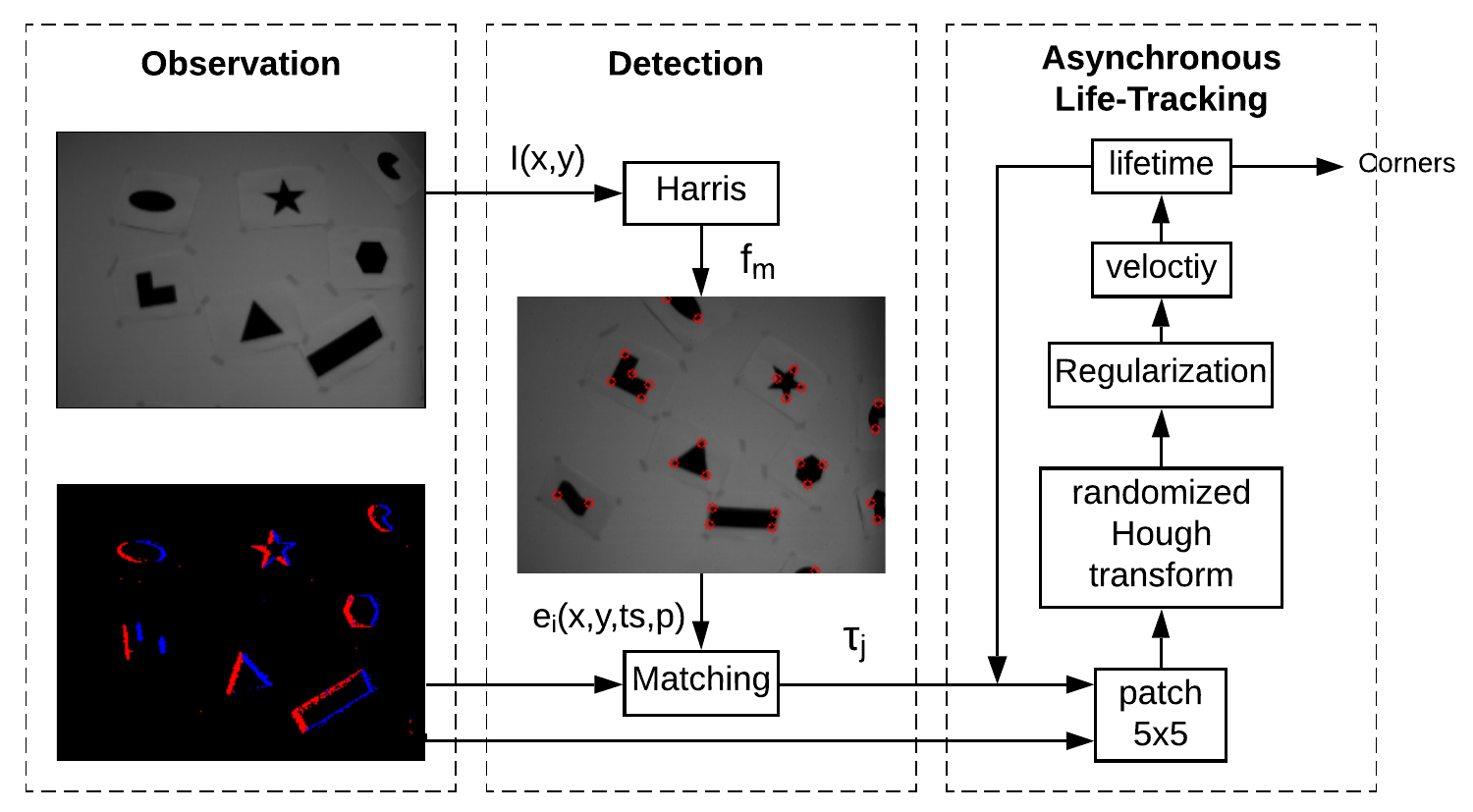}
    \caption{Overview of the overall asynchronous corner detection and tracking algorithm.}
    \label{fig:sys}
\end{figure}

The proposed approach, illustrated in Figure~\ref{fig:sys}, is composed of three main phases: observation, detection, and asynchronous tracking. Details of the various phases are discussed in the following subsections.

{}

\subsection{Observation}
The sensor of DAVIS \cite{davis} cameras has an Active Pixel Sensor (APS)~\cite{APS} and a Dynamic Vision Sensor (DVS) for each pixel, making them able to generate synchronized images and events at the same time. Generated images contain the absolute brightness of the scene and have a resolution of 240x180 pixels. Grayscale images are normally captured at a constant rate (equal to 24Hz). In our approach, we use images as a \textit{keyframe} to extract strong and robust corners. The other output of the DAVIS camera is events. Events represent the change of brightness of each pixel independently. Events are triggered asynchronously at a high rate (up to 1MHz), which makes them ideal for tracking in a highly dynamic environment. Events contain simple information: the position of the pixel, the timestamp, and the polarity of the brightness changes [ 1, -1]. 

\subsection{Detection}
\begin{figure}[ht]
\centering
\subfigure[corner region]{\includegraphics[width = 1.5in]{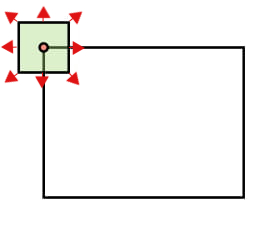}\label{fig:c1}}
\hspace{2pt}
\subfigure[edge region]{\includegraphics[width = 1.5in]{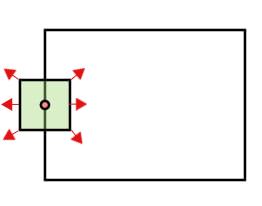}\label{fig:c2}} \hspace{2pt}
\subfigure[flat region]{\includegraphics[width = 1.5in]{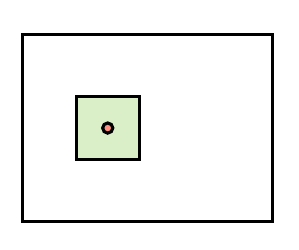}\label{fig:c3}}
\caption{A general presentation of three different regions in images: corners, edges, and flat regions. }
\label{fig:corner}
\end{figure}

In the detection phase, we extract 2-dimensional interest points, i.e., corners from the scene, using Harris \cite{harris} algorithm. As illustrated in Figure \ref{fig:c1}, a point in an image is considered as a corner when its local pixels have a significant change in all eight directions. Edges have significant changes in six direction but not along the edge direction Figure \ref{fig:c2}. On the other hand, flat regions have no change in all directions Figure \ref{fig:c3}. Thus, corners are important for location and object detection algorithm based in vision sensors, since they are invariant to motion changes, such as rotational and translation and illumination changes. Harris is considered one of the most robust corner detection algorithm in the literature. 

Harris algorithm uses a score technique to extract strong corners from the capture images. A point is considered as a corner if its score is larger than a certain threshold. A window size of 3x3 pixels is used to calculate the $S$ score of each pixel in the image, as follows

\begin{equation}
    \begin{split}
        &S = \lambda_{1} \lambda_{2}- k(\lambda_{1}+ \lambda_{2})^2 \\
    \end{split}
\end{equation}
where $\lambda_{1}$ and $\lambda_{2}$ are the eigenvalues of $M$, i.e.,
\begin{equation}
    M = \sum{w(x,y) \begin{bmatrix}I_{x}^{2}  & I_{x}I_{y} \\ I_{x}I_{y} & I_{y}^{2}\end{bmatrix}}
\end{equation}{}

where $w(x,y)$ denotes the local window of each pixel. The horizontal and vertical gradients is denoted by $I_{x}$ and $I_{y}$ respectively. Point $P$ is considered as a corner if its score $S$ is large, in other words its $\lambda_{1}$ and $\lambda_{2}$ are large and $\lambda_{1} \sim \lambda_{2}$. After we detect corners from grayscale images, we use a matching algorithm to extract event-corners. 

\begin{algorithm}[h]
\textbf{Input:} frame-corners, events\\
\textbf{Output:} event-corners 
\begin{algorithmic}[1]
\STATE Initialization
\IF{e.position = corner.position}
    \STATE Select L\_SAE
    \STATE Determine binary L\_SAE 
    \STATE Compute $I_{x}$ and $I_{y}$ 
    \STATE Compute Score $S$ 
    \IF{ $S$ $>$ threshold}
        \STATE Process event 
    \ELSE
        \STATE Discard event
    \ENDIF
\ELSE
    \STATE Discard event
\ENDIF
\end{algorithmic}
\caption{Matching Unit}
\label{algo:match}
\end{algorithm}

\begin{figure}[t]
\centering
\subfigure[local SAE]{\includegraphics[width = 1.5in]{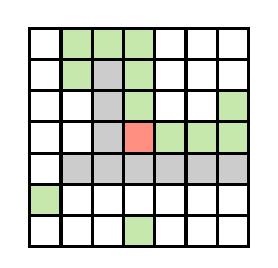}\label{fig:SAE1}}
\hspace{2pt}
\subfigure[binary SAE]{\includegraphics[width = 1.45in]{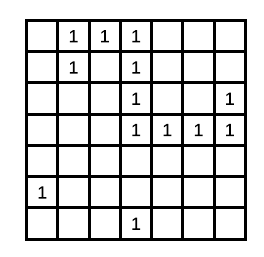}\label{fig:SAE2}} \hspace{2pt}
\caption{A presentation of local and binary SAE. Incoming event, i.e., corner-candidate, is in the center of the SAE in red. The most recent N neighbours are in green and are labeled as ``1'' in the binary SAE. Old neighbouring events are in gray and they are not included in the binary SAE.  }
\label{fig:b_SAE}
\end{figure}

The matching algorithm, as summarized in Algorithm \ref{algo:match}, filters the incoming event and only processes the first corner-candidate that occurs on the same location of the extracted corners. Processed events, i.e., event-corners, are then fed to the next phase to track the event-corners in asynchronous fashion using neighbouring events. The inputs of the algorithm are the triggered events and the extracted frame-corners from Harris. If the incoming event is on the same location of a corner, a local surface active events (L\_SAE) is extracted around the event. The L\_SAE has a size of 7x7 pixels and contains the timestamp on neighboring events. As shown in Figure \ref{fig:b_SAE}, only the most recent $N$ neighbours  ($N$  =  12)  are  included  and  labeled  as 
``1'' in binary L\_SAE to compute the vertical and horizontal gradients. Those gradient are then used to compute the score of the incoming event. If the score is larger than the threshold, the incoming event is considered as event-corner and it's only updated when a new image, i.e., keyframe is captured.

\subsection{Asynchronous Life-Tracking}

In this section, we explain our asynchronous tracking algorithm which is based on the lifetime of events. The concept of lifetime refers to the time an event will take to move from the current pixel to one of the eight neighboring pixels. Since event cameras have a high temporal rate up to 8 million events per second, lifetime is feasible for event-based tracking. The summary of the asynchronous tracking algorithm is illustrated in Algorithm \ref{alg:aTracking}, in which we first extract a local L\_SAE of size 5x5 pixels around each event-corner. The randomized Hough transform algorithm \cite{RHT} is used to fit a local plane using the neighboring events in the L\_SAE. The vector ($a$,$b$, and $c$) which is orthogonal onto the obtained plane is calculated. Then we compute the velocity of the event-corner in x- and y-directions $v_x$ and $v_y$ respectively , as follows:

\begin{equation}
    v_x = c * (-a) / (a^2+b^2)
\end{equation}

\begin{equation}
    v_y= c * (-b) / (a^2+b^2)
\end{equation}

The plane fitting algorithm is inspired by \cite{lifetime}, in which we robustly compute candidate planes using three points, i.e., the event-corner and two additional neighboring events from L\_SAE. The plane fitting algorithm is illustrated in Algorithm \ref{alg:RHT}, where two neighboring are chosen randomly plus the event-corner to fit a local plane (Line 1). The algorithm calculated the Hough space parameters H($\theta, \phi, \rho$) and store in different spaces in cells $C$ (Lines 3-4). For each vote, a cell counter $C$ increments by one  $C + 1$ (Lines 5-6) and the most voted cell is considered as the fitting plane. The Hough space is computed using two randomly picked points $p_{2}$,$p_{3}$ from the point set in L\_SAE and the event-corner $p_{1}$, i.e.,

\begin{algorithm}[t]
\textbf{Input:} \textbf{$e_i$ = ($x_i$, $y_i$, $t_i$, $pol_i$)} and $\tau = \{e_1, e_2, .., e_j\}$\\
\textbf{Output:} new location of events in $\tau$
\begin{algorithmic}[1]
\STATE Extract SAE of size 5x5 for each event in $\tau$\;
\STATE Fit a local plane for each SAE\;
\STATE Calculate $a_j$,$b_j$, and $c_j$\;
\STATE Calculate velocity $v_j(v_x, v_y)$\;
\STATE Calculate lifetime\;
\end{algorithmic}
\caption{Asynchronous Life-Tracking}
\label{alg:aTracking}
\end{algorithm}
\vspace{-0.5cm}

\begin{equation}
    \rho = v\cdot p_1=((p_3-p_1)\times(p_1-p_2))\cdot p_1
\end{equation}

\begin{equation}
    p_x \cdot \cos{\theta} \cdot \sin{\phi} + p_y \cdot sin{\phi} \cdot \sin{\theta} + p_z \cdot \cos{\phi} = \rho
\end{equation}

where the distance to the centroid is denoted by $\rho$,  $\phi$ is the angle between the vector $v$ and the XY-plane in the z-direction and $\theta$ denotes the angle of $v$ on the XY-plane. 

\begin{algorithm}
\textbf{Input:} \textbf{P = \{$p_{1}$,$p_{2}$, .., $p_{n}$\}}\\
\textbf{Output:} H($\theta, \phi, \rho$)
\begin{algorithmic}[1]
\WHILE{points in \textbf{P} $> 2$}
    \STATE Select event-corner \textbf{$p_{1}$} and two additional points \textbf{$p_{2}$}, \textbf{$p_{3}$}
    \STATE Calculate Hough space
    \STATE Store planes in cells
    \STATE Points vote for cells
    \STATE For each vote cell's Counter C = C + 1
    \IF{C = threshold}
        \STATE Parameterize the detected plane
        \STATE Delete $p_2$ and $p_3$ from \textbf{P}
        \STATE C = 0
    \ELSE
        \STATE Continue
    \ENDIF
\ENDWHILE
\end{algorithmic}
 \caption{Randomized Hough transform}
 \label{alg:RHT}
\end{algorithm}
\vspace{-1cm}

\section{Experimental Evaluation}
We evaluated the proposed approach on the publicly available datasets~\cite{Dataset}. Subsets used in the experiment consists of slow and fast camera motions and low- and high-textured scenes to ensure comprehensive evaluation scheme. In general, DAVIS camera can generate up to 10 million events per second in textured and highly dynamic scenes. On other hand, in slow camera motion and texture-less scene, event camera triggers few events. As mentioned previously, DAVIS camera produce a sequence of intensity images along side a stream of asynchronous events. The subsets used in the experiment are recorded by a DAVIS240 camera which generates images with resolution of 240x180. We implemented our algorithm in C++, OpenCV and Eigen libraries. All experiments were carried out on an embedded system with an ARM-based processor, hosted on Jetson TX2 board and running at 2 GHz.

\begin{figure}[t]
\centering
\subfigure[frame-corners]{\includegraphics[width = 1.4in]{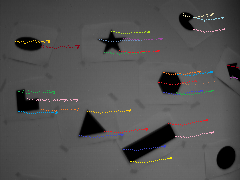}\label{fig:r1}}
\hspace{2pt}
\subfigure[frame-corners in slow motion]{\includegraphics[width = 1.4in]{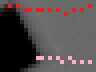}\label{fig:r2}} 
\hspace{2pt}
\subfigure[frame-corners in rapid motion]{\includegraphics[width = 1.68in]{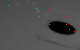}\label{fig:r3}} 

\subfigure[event-corners]{\includegraphics[width = 1.4in]{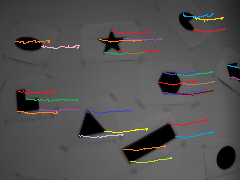}\label{fig:r4}}
\hspace{2pt}
\subfigure[event-corners in slow motion]{\includegraphics[width = 1.4in]{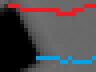}\label{fig:r5}} 
\hspace{2pt}
\subfigure[event-corners in rapid motion]{\includegraphics[width = 1.68in]{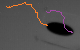}\label{fig:r6}} 
\caption{The qualitative result of proposed method compared with Harris method on intensity images.}
\label{fig:result}
\end{figure}

The experimental results of the proposed asynchronous corner tracking algorithm are reported in Figure \ref{fig:result}. For this evaluation we used \textit{shapes\_6dof} dataset. The dataset includes various 2D geometric shapes mounted on a wall and recorded freely moving DAVIS-240 cameras with different speeds. We compared the proposed algorithm with normal Harris corner detection and at different speeds of the camera. In slow camera motions, frame-corner detection method, i.e., Harris performances considerably well as only break of 1 pixel appears in the detection, as shown in Figure \ref{fig:r2}. However, when the speed of the camera, or some objects in the environment, is high, feature extraction and tracking based on intensity images cannot provide sufficient information about the environment. This can be shown in Figure \ref{fig:r3} where the increase in the speed of the camera imposes a break in detecting up to 4 pixels between two consecutive detections. In such an environment, an event-based camera and the proposed feature tracking technique can provide a seamless feature tracking for the algorithm, see Figure \ref{fig:r6}. 

Table \ref{table:table1} summarizes the execution time of different units in the proposed method. The intensity images have a resolution of 240x180, which leads to fast execution time for Harris algorithm, i.e., the average execution time is 1.8 milliseconds. Execution time for Harris does not change significantly since the algorithm processes all pixels in the image and all captured images have the same size. On the other hand, event-based units, such as matching unit and lifetime calculation depends on the number of generated events that is based on the speed of the camera  and the amount of information in the scene \cite{sherif1}. The execution time of both units is very low and especially the matching unit which has an average execution time of 3.6 microseconds. 
One of the main advantages of the proposed solution w.r.t. the state-of-the-art event-based approaches is the fact that it is able to provide high performance in all the scenarios; in particular, in case the camera moves very fast, up to 10 Millions events can be generated. This tremendous load causes an execution time of the classical  matching unit equal to 36 seconds and of the lifetime calculation equal to 4900 seconds, thus not affordable in any real-time scenario. At the opposite, the proposed filtering solutions has an average execution time of 5.62 seconds. We extract a maximum of 50 corners per image using Harris algorithm from images. The DAVIS-240 camera captures a total number of 24 images per second. For each corner an average number of 10 events are checked to detect event-corners using the matching unit, which totals a number of 12$\times10^3$ events are checked per second resulting a maximum execution time of 43.2 milliseconds.

\newcommand{\centered}[1]{\begin{tabular}{l} #1 \end{tabular}}
\begin{table}
\caption{The average execution time of units of the proposed method} 
\centering
\begin{tabular}{|l|c|} 
\hline
\centered{Unit} & \centered{Time (ms)}\\
\hline\hline
\centered{Harris per image} & \centered{1.8} \\ 
\hline
\centered{matching per event} & \centered{3.6$\times10^{-3}$} \\
\hline
\centered{lifetime per event} & \centered{0.49} \\ 
\hline
\end{tabular}
\label{table:table1}
\end{table}
\vspace{-1cm}
\section{Conclusion}

Performing high-frequency corner detection and tracking based on traditional cameras is not feasible due to the limitations of such cameras, such as motions blur and blind time. In this paper, we exploit event-based cameras since they can cope well with rapid camera movements and highly dynamic scenes. We proposed an asynchronous tracking algorithm based on the lifetime of extracted event-corners. We first extract strong frame-corners from intensity images using Harris. Afterward, we match those corners with first generated events with the same pixel location. The tracking algorithm fits a local plane to the neighboring events of each extracted event-corner to compute the lifetime and velocity of each event-corner. A randomize Hough transform is used to fit a plane. Finally, we evaluated our proposed method by performing experiments on datasets with different speeds and compared the results with the extracted corner using only intensity images. In addition, the execution time of each unit of the proposed method is reported. The results show that our method has an average execution time of 5.62 seconds.

\bibliographystyle{splncs04}

\bibliography{refs}
\end{document}